%% file: emnlp2020.tex
\documentclass[11pt,a4paper]{article}
\usepackage{authblk}
\usepackage[hyperref]{emnlp2020}
\usepackage{times}
\usepackage{latexsym}

\usepackage{amsmath}
\usepackage{amssymb}

\usepackage{url}
\usepackage{booktabs} 
\usepackage{graphicx}
\usepackage{epstopdf}
\usepackage{subfigure}
\usepackage{verbatim}
\usepackage{amssymb}
\usepackage{amsmath}
\usepackage{bm}
\usepackage{array}
\usepackage{multirow}
\usepackage{subfigure}
\usepackage[most]{tcolorbox}

\usepackage[font={small}]{caption}

\usepackage{microtype}

\aclfinalcopy 

\title{Logic2Text: High-Fidelity Natural Language Generation\\ from Logical Forms}

\author[1]{\textbf{Zhiyu Chen}}
\author[1]{\textbf{Wenhu Chen}}
\author[1]{\textbf{Hanwen Zha}}
\author[1]{\textbf{Xiyou Zhou}}
\author[1]{\textbf{Yunkai Zhang}}
\author[2]{\\\textbf{Sairam Sundaresan}}
\author[1]{\textbf{William Yang Wang}}
\affil[1]{University of California, Santa Barbara}
\affil[2]{Intel AI \authorcr \{zhiyuchen, wenhuchen, hwzha, xiyou, yunkai\_zhang, william\}@cs.ucsb.edu, \authorcr sairam.sundaresan@intel.com}
\date{}

\begin{document}
\maketitle

\input{00-abstract.tex}
\input{01-introduction.tex}

\input{02-related.tex}

\input{03-dataset.tex}
\input{04-dataset_stats.tex}
\input{05-experiments.tex}
\input{06-conclusion.tex}

\section*{Acknowledgment}
We thank the anonymous reviewers for their thoughtful comments.
This research was sponsored in part by Intel AI Faculty Research Grant and NSF IIS 1528175. The authors are solely responsible for the contents of the paper and the opinions expressed in this publication do not reflect those of the funding agencies.

\bibliography{emnlp2020}
\bibliographystyle{acl_natbib}

\input{07-appendix.tex}

\end{document}

%% file: 00-abstract.tex
\begin{abstract}
Previous studies on Natural Language Generation (NLG) from structured data have primarily focused on surface-level descriptions of record sequences. However, for complex structured data, e.g., multi-row tables, it is often desirable for an NLG system to describe interesting facts from logical inferences across records. If only provided with the table, it is hard for existing models to produce controllable and high-fidelity logical generations. In this work, we formulate high-fidelity NLG as generation from logical forms in order to obtain controllable and faithful generations. We present a new large-scale dataset, \textsc{Logic2Text}, with 10,753 descriptions involving common logic types paired with the underlying logical forms. The logical forms show diversified graph structure of free schema, which pose great challenges on the model's ability to understand the semantics. We experiment on (1) Fully-supervised training with the full datasets, and (2) Few-shot setting, provided with hundreds of paired examples; 
We compare several popular generation models and analyze their performances.
We hope our dataset can encourage research towards building an advanced NLG system capable of natural, faithful, and human-like generation. The dataset and code is available at \url{https://github.com/czyssrs/Logic2Text}.
\end{abstract}

%% file: 01-introduction.tex
\section{Introduction}
\begin{figure}[ht]
\centering
\includegraphics[width=0.48\textwidth]{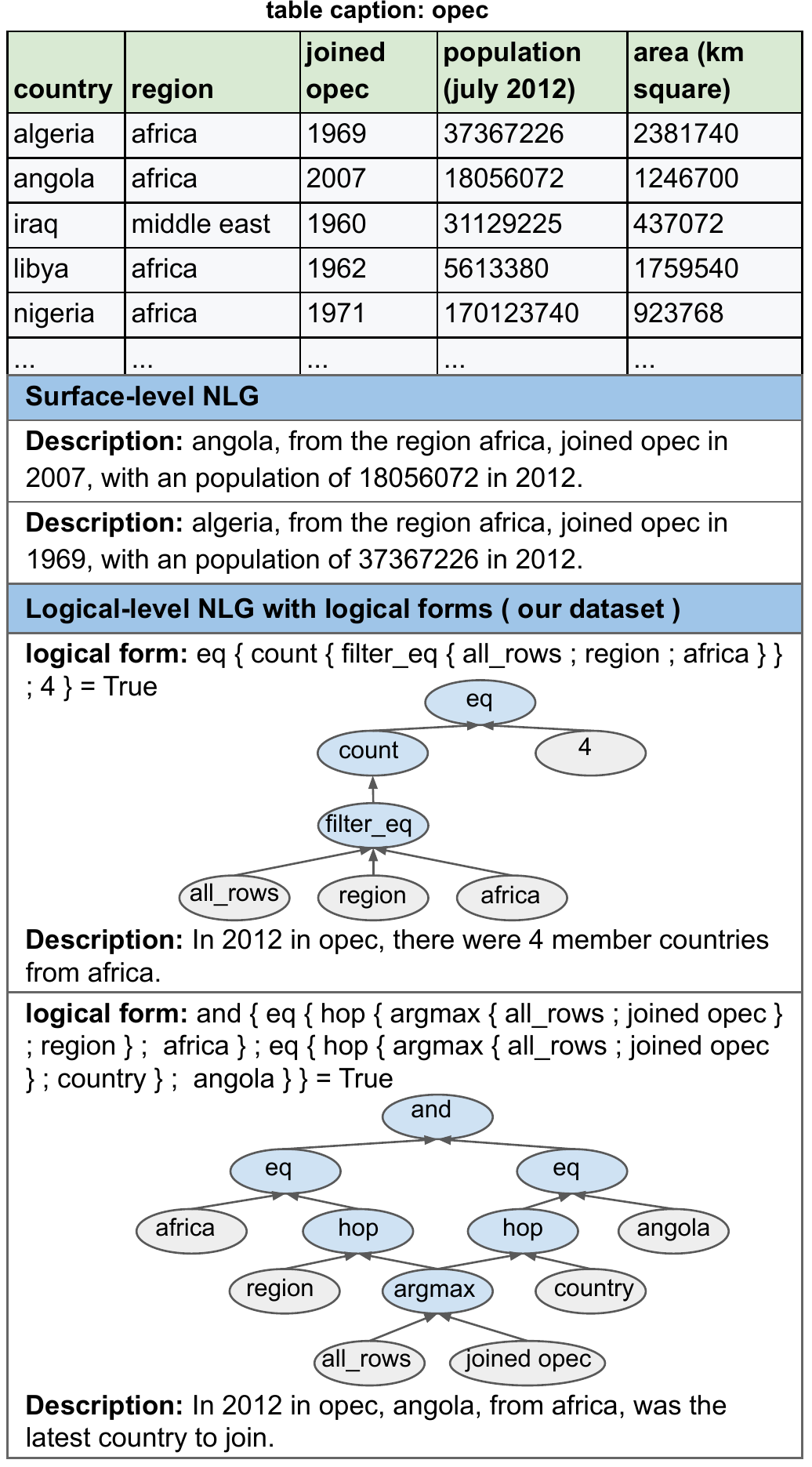}
\caption{Examples of surface-level NLG compared with NLG with logical forms of our dataset. Here are two examples with logic type \texttt{count} and \texttt{superlative}. The function nodes are in blue, and the text nodes in grey.} 
\label{fig:eg0}
\end{figure}

Natural language generation (NLG) from structured data has been an important research problem in many applications.
Recent data-driven methods have achieved good performances on various NLG tasks~\cite{DBLP:conf/aaai/LiuWSCS18, DBLP:conf/emnlp/FreitagR18, DBLP:journals/corr/abs-1904-09521}. However most studies focus on surface descriptions of simple record sequences, for example, attribute-value pairs of fixed or very limited schema, like E2E~\cite{DBLP:conf/sigdial/NovikovaDR17} and WikiBio~\cite{DBLP:conf/emnlp/LebretGA16}.
In real-world cases for multi-row tables, it is often more desirable and plausible to provide descriptions involving higher-level logical inference across data records. For example, in Figure~\ref{fig:eg0}, instead of plain restatements, human readers would be more favorable to abstract descriptions that can summarize or conclude information over the table records. 
To produce such logical-level generations of high fidelity, it is not yet appropriate to provide only the table as the input in a real-world NLG system, based on the following reasons:

1) \textit{Low Fidelity}. Given only the table, it is challenging for existing neural models to produce such logically correct generations involving reasoning and symbolic calculations, e.g., \texttt{max}, \texttt{min}, \texttt{counting}, \texttt{averaging}, etc. 

2) \textit{Uncontrollable content selection}. Given a table, the space of logically entailed descriptions is exponentially large, due to vast number of combinations of different operations and arguments from the table, e.g., \texttt{count}, \texttt{comparison}, \texttt{superlative}, etc. 
It is hard and uncontrollable for neural models to decide a valid, favorable choice of logical selections solely based on the table, due to the difficulty of imposing high-level semantic constraints in the compositional generation process. 

To combat with the above problems, we argue that it is necessary to leverage intermediate meaning representations to achieve faithful and controllable logical generations. 
To this end, we formulate the task of logical-level NLG as a \textbf{logical form to text} problem. Specifically, besides the table information, the generation module is provided with a logical form representing the semantics of the target text (see Figure~\ref{fig:eg0} for an example). 
By separating logical reasoning and language realization, the correctness of the intermediate logical form is guaranteed, and the challenge for the realization module is fully shifted to semantic understanding.

To facilitate research in this direction, we propose a new dataset named \textsc{Logic2Text}, consisting of 5.6k open-domain tables, 10.8k manually annotated (logical form, description) pairs. 
Our dataset is of high quality in terms of (1) natural and interesting descriptions; (2) accurate logical forms with 100\% execution correctness. In our dataset, the coarse logic types are 7 common ones to describe multi-row tables: \texttt{count}, \texttt{superlative}, \texttt{comparative}, \texttt{aggregation}, \texttt{majority}, \texttt{unique}, and \texttt{ordinal}. 
We employ a Python-like program to serve as our logical forms, which can be easily converted to other types of logical forms. 
Figure~\ref{fig:eg0} shows two examples of our dataset. Compared with previous surface-level NLG datasets, one major distinction of our dataset is the free schema of the logical forms, which can be represented as diversified graph structures. The new dataset poses great challenges on the model's ability to understand the structural semantics in graph representation. 

We employ an array of popular generation models as the baseline approaches. The experiments are conducted in (1) \textit{Fully-supervised setting.} We train the models using the full dataset to analyze their performances. (2) \textit{Few-shot setting.} We simulate the low-resource scenario in real-world use cases.
Experimental results show that the logical forms are critical to acquiring high-fidelity generations. The pre-trained language model outperforms other baselines (pointer-generator, graph2seq, transformer, etc.), but still makes factual and logical errors. 

In summary, our contributions are the following: 
\begin{itemize}
    \item We propose a new large-scale dataset, \textsc{Logic2Text}, with descriptions of common logic types accompanied by the underlying logical forms. The logical forms present diversified graph structures, which raises more challenges on semantic understandings.
    \item We surveyed several popular generation models as the baselines under fully-supervised and few-shot settings, as well as analyze their pros and cons. 
\end{itemize}

Our dataset can also be used in the reverse way (text to logical form) to facilitate tasks related to semantic parsing. \citet{DBLP:journals/corr/abs-1909-02164} propose the task of fact verification against tables, however the performance is greatly limited due to the lack of the ground truth logical forms. This can be one direct application of our dataset. In this work, we focus on NLG. 


%% file: 02-related.tex
\section{Related Work}
NLG from structured data or knowledge has been studied for many years. There are various applications, such as the automatic generations of weather reports~\cite{DBLP:conf/acl/LiangJK09}, sport reports~\cite{DBLP:conf/emnlp/WisemanSR17}, clinical and health reports~\cite{dimarco2007development, lee2018natural}, response generation in task-oriented dialogue systems~\cite{DBLP:conf/emnlp/WenGMSVY15,DBLP:conf/emnlp/BudzianowskiWTC18,dusek2019e2e}, etc.

Traditional methods typically employ a pipeline-based approach including content selection, planning and surface realization~\cite{DBLP:journals/nle/ReiterD97, DBLP:journals/jair/GattK18}. Recent data-driven methods tend to conflate the pipeline modules into one end-to-end neural networks, such as~\cite{DBLP:conf/aaai/LiuWSCS18,DBLP:conf/emnlp/WisemanSR17, DBLP:conf/emnlp/WisemanSR18, DBLP:conf/emnlp/GongFQL19}.
Most recently, large-scale pre-trained models~\cite{radford2019language, DBLP:conf/icml/SongTQLL19, DBLP:journals/corr/abs-1910-10683} have achieved new state-of-the-arts on various generation tasks. 
Chen et al.~\shortcite{DBLP:journals/corr/abs-1904-09521} demonstrate that a simple pre-training based method can achieve very reasonable performance on the WikiBio dataset~\cite{DBLP:conf/emnlp/LebretGA16} under few-shot setting. 
More recent works begin to focus on fidelity preserving of the generation, such as~\cite{DBLP:conf/acl/DhingraFPCDC19,DBLP:journals/corr/abs-1910-08684}. Their work obtains good performances on surface-level NLG. In contrast, our work focus on the fidelity of logical-level generations. 

There are a few popular NLG datasets mostly on surface-level generation. Such as WeatherGov~\cite{DBLP:conf/acl/LiangJK09}, E2E~\cite{DBLP:conf/sigdial/NovikovaDR17}, WikiBio~\cite{DBLP:conf/emnlp/LebretGA16}, and ToTTo~\cite{DBLP:journals/corr/abs-2004-14373}. RotoWire~\cite{DBLP:conf/emnlp/WisemanSR17} is a more challenging dataset on generating basketball game reports from multi-row tables. But the reports are still limited to superficial restatements of table records, with very few involving logical inference. \citet{DBLP:conf/sigmod/KornWWY19} investigate generation of interesting trivia from superlative wikipedia tables. 
\citet{chen2020logic} propose the task of generating arbitrary sentences with logical inference from the table. Their task mainly works for probing purpose, i.e., to test the ability of neural models to produce any logically correct descriptions solely based on the table. However, such a task formulation is not yet appropriate for building a real-world NLG system due to low-fidelity, as we discussed in the introduction. The best-performing model in~\cite{chen2020logic} only obtains a factual correctness rate over 20\% based on human evaluation, which is clearly far from an acceptable level in real-world systems. 

Another line of works related to ours is the text generation from syntactic or semantic sentence structure, such as generation from CCG grammar~\cite{white2006efficient}, UCG grammar~\cite{gardent1990generating}, AMR~\cite{DBLP:conf/acl/GildeaWZS18}. There are many early works attempting algorithmic approaches on such kinds of logical formulations~\cite{DBLP:journals/mt/Phillips93, DBLP:conf/eacl/CalderRZ89, DBLP:journals/coling/ShieberNPM90, DBLP:journals/mt/Phillips93}, etc. Later proposed datasets include the Groningen Meaning Bank~\cite{DBLP:conf/acl-jssp/Bos13}, the AMR bank~\cite{DBLP:conf/semeval/May16}, the DeepBank~\cite{flickinger2012deepbank}, etc. In contrast, our work focus on the logical formulations executed on database style tables, and common symbolic operations on tables, such as count, superlative, comparison. As nowadays much of the production data is stored in table based DB, we believe such a dataset should help building systems with table based data.

%% file: 03-dataset.tex
\section{Dataset Construction}

\begin{figure*}[ht]
\centering
\includegraphics[width=0.96\textwidth]{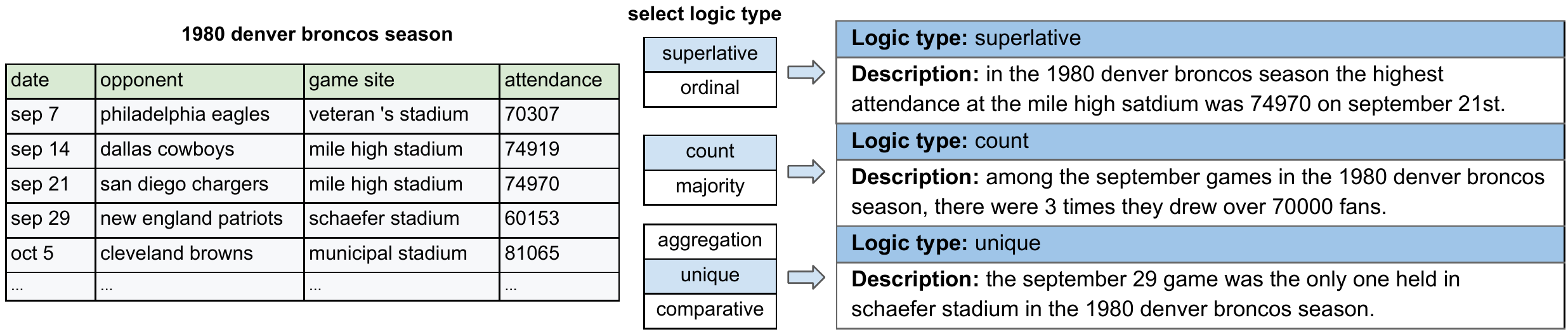}
\caption{description composition: the workers are asked to select three logic types and compose a statement based on the selected logic type, that describe interesting facts in the table.} 
\label{fig:step1}
\end{figure*}

The table source of \textsc{Logic2Text} is from WikiTables\footnote{\url{http://websail-fe.cs.northwestern.edu/wikiTables/about/}}~\cite{DBLP:conf/kdd/BhagavatulaND13}, a collection of open-domain tables crawled from Wikipedia. We follow~\cite{DBLP:journals/corr/abs-1909-02164} to filter out over-complicated tables and take a subset of tables with less than 20 rows and 10 columns.

In this dataset, we start from 7 types of most commonly used logics~\cite{DBLP:journals/corr/abs-1909-02164} to describe multi-row tables: \texttt{count}, \texttt{superlative}, \texttt{comparative}, \texttt{aggregation}, \texttt{majority}, \texttt{unique}, and \texttt{ordinal}. For example, for logic type \texttt{count}, the definition is: counting some rows in the table based on the values in one column, with the scope of all table rows or a subset. Refer to Appendix A for the definitions of all logic types. Each description involves exactly one type of logic. This matches the observation that humans generally do not describe their interested information in tables with over-complicated logics. For logical forms, we use a python-like program, and the function set is an extension of~\cite{DBLP:journals/corr/abs-1909-02164}. Refer to Appendix B for definitions of all functions. 

Our dataset is constructed in 3 stages: ~\S\ref{data:step1} Description composition and verification, ~\S\ref{data:step3} Logical form annotation and derivation, ~\S\ref{data:step4} Logical form execution and verification. 
We adopt the workflow of composing descriptions first and then deriving the logical forms, because under such an order, the annotators can compose natural descriptions based on the interesting facts in the table, which is hard to be achieved by automatic enumeration of logical forms followed by template re-writing. 
For all crowd-sourcing tasks we hire Amazon Mechanical Turkers\footnote{\url{https://www.mturk.com/}} (AMT) under three requirements: (1) from English native countries (``US",``CA",``GB", ``AU"); (2) Approval rate higher than 95\% for all HITs; (3) More than 500 approved HITs. We follow the human subject research protocols\footnote{\url{https://en.wikipedia.org/wiki/Minimum\_wage\_in\_the\_United\_States}} to pay the workers. We maintain strict high criterions for approval and review at least 10 random samples for each worker to decide whether to approve or reject all his/her HITs.

\subsection{Description Composition \& Verification}
\label{data:step1}
In this first stage, the human workers are asked to compose statements of a \textit{certain logic type}, that describe \textit{interesting} facts in the table. It's possible that some logic types cannot be applied to certain tables. Therefore we design the following working procedure: For each table, the 7 logic types are randomly put into three groups (with sizes 2, 2, and 3). The worker is asked to choose one logic type from each group and compose a description based on the chosen logic type. They must follow the requirements (1) try to choose diversified logic types, (2) avoid template-like language and try to compose natural and interesting descriptions, (3) include the information in table captions, so as to compose comprehensive descriptions without unspecified pronouns. 
An example of the workflow is shown in Figure~\ref{fig:step1}. We provide the workers detailed explanations for each logic type by their corresponding definitions, accompanied by examples. 
After collecting the descriptions, we add a verification stage to filter out descriptions of low quality. We redistribute the collected descriptions grouped by each logic type, then ask three questions: Is this description (1) of the correct logic type presented? (2) factually correct? (3) grammatically correct and fluent? We filter out the description if any question receives a negative response. 

\subsection{Logical Form Annotation \& Derivation}
\label{data:step3}
\begin{figure*}[ht]
\centering
\includegraphics[width=1.0\textwidth]{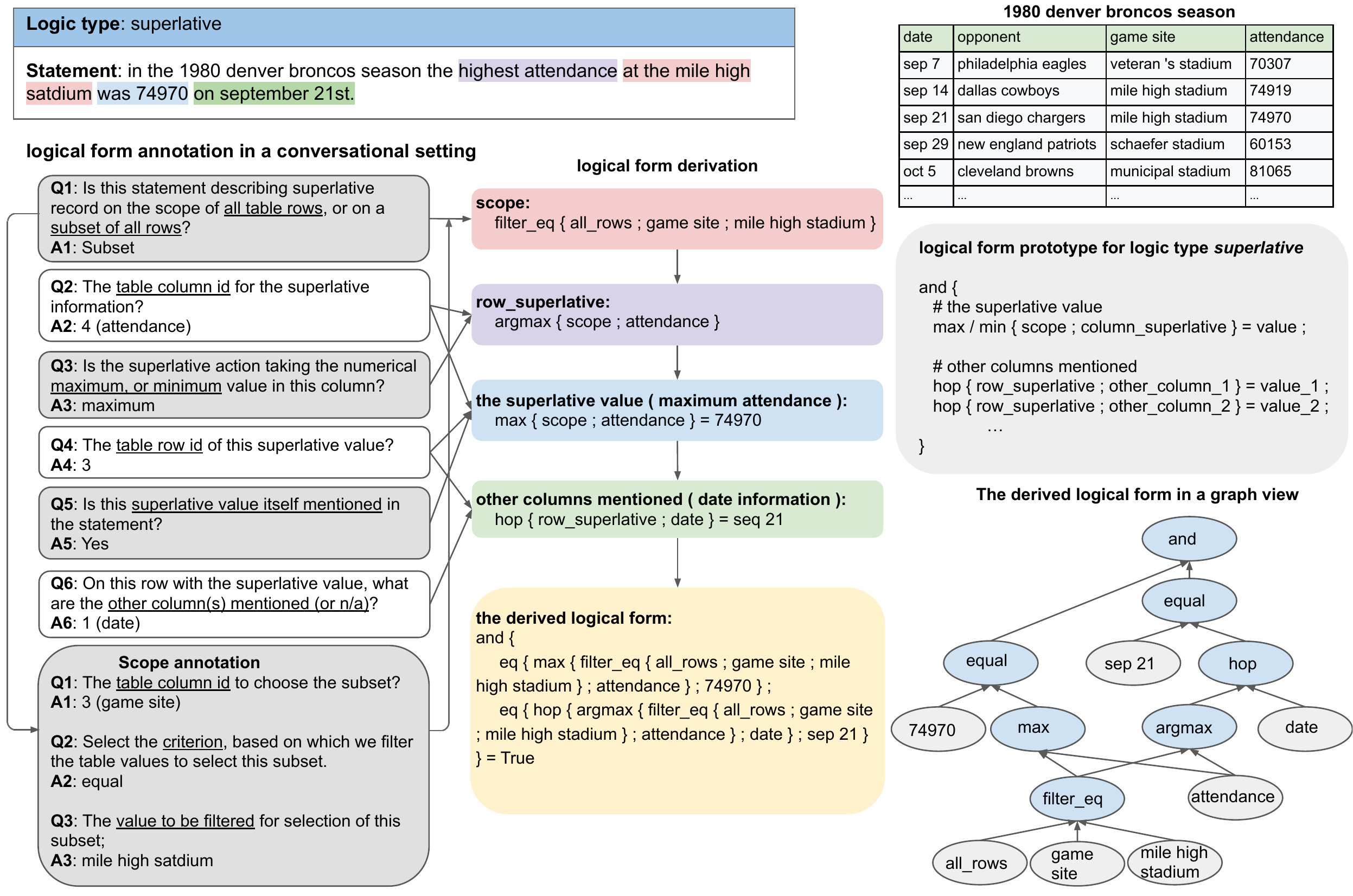}
\caption{logical form annotation \& derivation: Note that in this example the questions are all in concise forms. In the AMT interface shown to the workers, we write instructions in a more casual and detailed manner, accompanied by several examples.} 
\label{fig:step3}
\end{figure*}

As the core step of our dataset construction pipeline, we design a workflow to obtain the semantic information via conversations with human workers, then use the information to derive the logical forms. The questions in the conversation are specifically designed for each logic type. 
Here we go through the example of logic type \texttt{superlative} given in Figure~\ref{fig:step3} to illustrate our annotation process. 

The logical form structure prototype is shown in the right grey part, consisting the description of the superlative value, and other mentioned columns on the row with the superlative value. 
Then we ask the follow-up questions to derive the complete logical form based on the prototype, shown on the left part of Figure~\ref{fig:step3}: Q1. What is the scope of the superlative operation? If the scope is a subset of all table rows, we perform another round of conversation to annotate the scope. Q2. What is the table column of the superlative operation? Q3. What is the specific type of the superlative operation: maximum or minimum. Q4. What is the table row with the superlative value. Q5. Is the superlative value itself mentioned in the description or not? Q6. What are the other columns mentioned in the description? After collecting the answers of the above questions, we can derive the logical form, as shown in the middle part of Figure~\ref{fig:step3}.

We provide the workers with detailed explanations of the prototype for each logical types, as well as several examples. Note that the prototype covers most, but not all of the logical descriptions due to their diverse nature. Thus we also provide the option to skip the example if it cannot be formulated by the given question set. Check Appendix A for the annotation process of other logic types.

\subsection{Logical Form Execution \& Verification }
\label{data:step4}
After the collection of logical forms, we use the Stanford CoreNLP toolkits\footnote{\url{https://stanfordnlp.github.io/CoreNLP/index.html}} to tokenize all text content (all table information, the descriptions, and the texts in the logical forms). 
To remove incorrect logical forms, we execute the logical forms and perform another round of semantic verification. 

\textbf{Logical Form Execution} The functionality in our logical form is based on the ones used in~\cite{DBLP:journals/corr/abs-1909-02164}. We extend the function set to deal with semi-structured table cells (dates, mixed numbers and strings, etc.). 
We execute all logical forms against the corresponding table, and only keeps the ones that evaluate to \texttt{True}. This guarantees that the logical forms in our dataset achieve 100\% execution correctness. 

\textbf{Semantic Verification}
Note that execution correctness does not guarantee semantic correctness. Therefore we perform another round of semantic verification. Since AMT workers do not have experts knowledge to understand the logical forms, we convert the logical form into natural language interpretation based on the operations of each function.  
We then ask the workers to verify whether the interpretation correctly matches the meaning of the description, with neither insufficient nor redundant information. Then we remove the examples receiving negative responses. 

\textbf{Expert Evaluation} To demonstrate the quality of our dataset, we employ two computer science graduate students to conduct evaluations. We randomly sample 200 examples for each logic type to verify the semantic correctness. Each example is examined by both students, and the decision is made after discussion. The result shows that each logic type reaches a correct rate no less than 90\%. 

%% file: 04-dataset_stats.tex
\begin{table}[htbp]
\small
\begin{center}
\resizebox{0.45\textwidth}{!}{%
\begin{tabular}{lr}
\toprule
Tables & 5,554\\
Examples & 10,753\\
Vocabulary & 14.0k\\
Avg. description length & 16.77\\
Avg. \# nodes in logical form & 9.00\\
Avg. \# function nodes in logical form & 3.27\\
Avg. length of the linearized logical form & 24.35\\
\bottomrule
\end{tabular}
}
\caption{General statistics of \textsc{Logic2Text}.}
\label{table:gen_stats}
\end{center}
\end{table}
\vspace{-10pt}
\begin{figure}[htbp]
\centering
\includegraphics[width=0.45\textwidth]{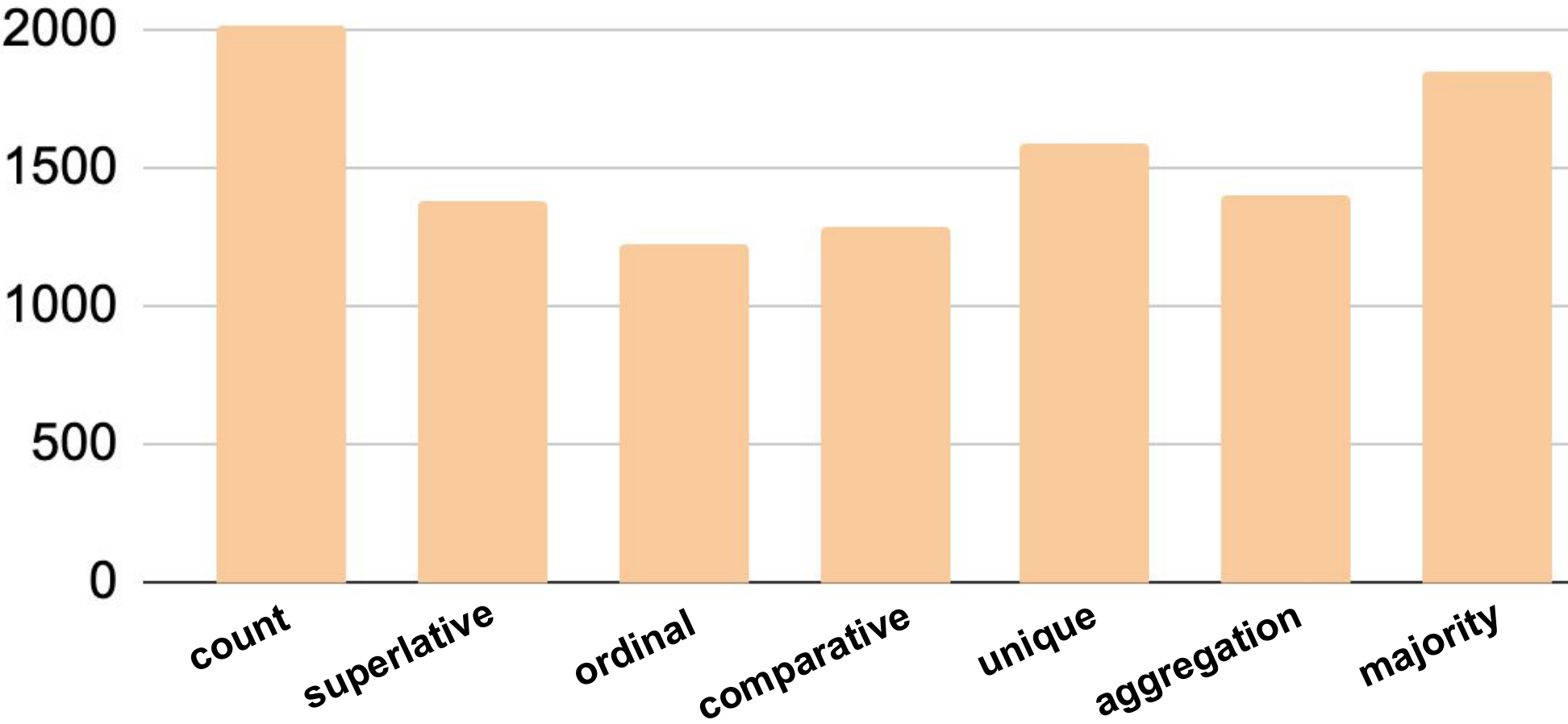}
\caption{Distribution of logic types.} 
\label{fig:logic_dist}
\end{figure}
\begin{figure*}[htbp]
\centering
\includegraphics[width=1.0\textwidth]{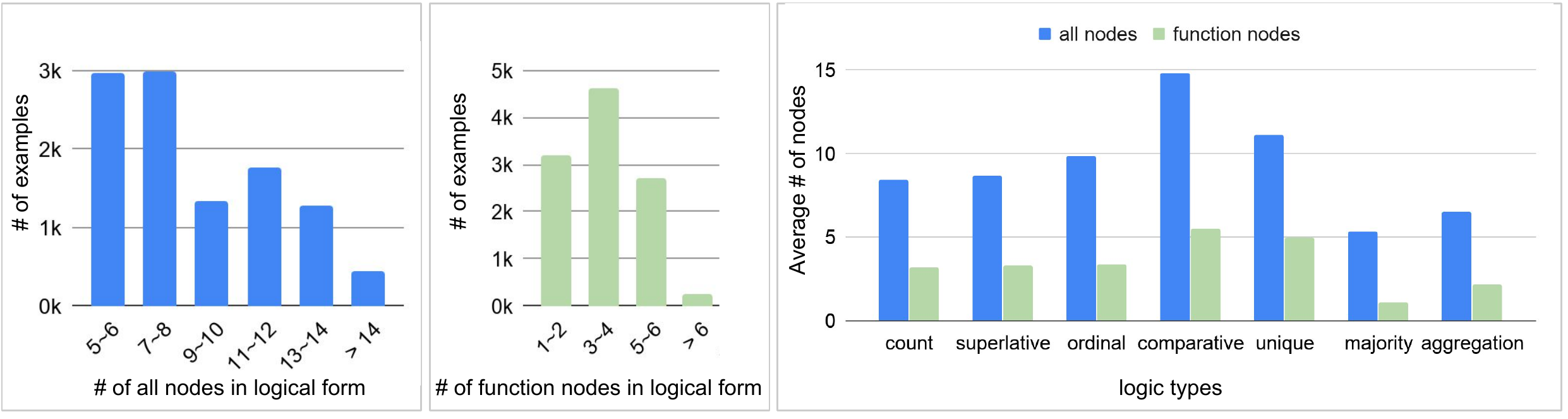}
\caption{The distribution of our dataset regarding the number of all nodes (\textit{Left}) and function nodes (\textit{Mid}) in the logical form. \textit{Right:} average number of all nodes and function nodes in the logical forms for each logic type. }
\label{fig:node_num}
\end{figure*}
\begin{figure*}[ht]
\centering
\includegraphics[width=1.0\textwidth]{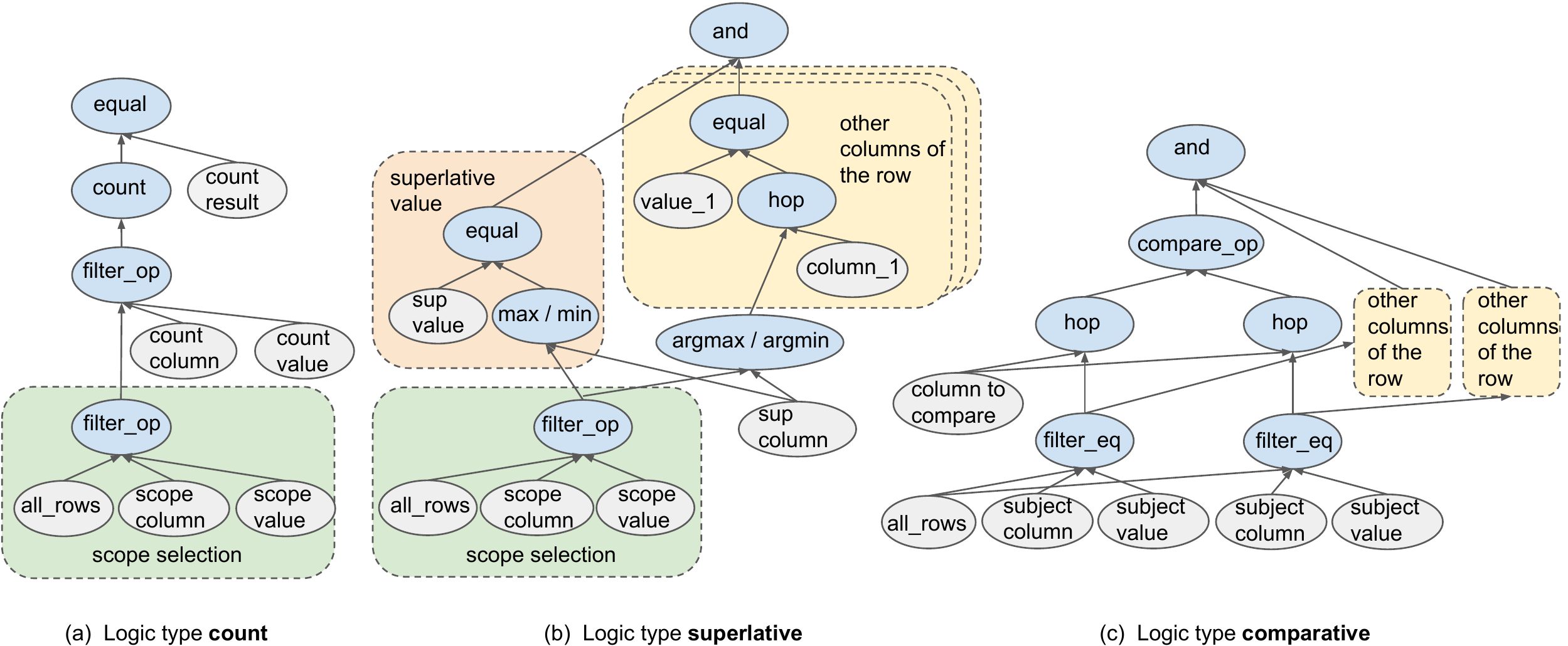}
\caption{Overview of logical form structures for logic type \texttt{count}, \texttt{superlative}, and \texttt{comparative}. (a) \texttt{count}: the structure in the green shadow is optional, representing the scope of counting. It can be all table rows (a single text node) or a subset of rows from a filter operation. (b) \texttt{superlative}: the structure in the orange shadow is optional, depending on the presence of the max/minimum value in the description. The structure in the yellow shadow appears 0 or more times.}
\label{fig:structure}
\end{figure*}
\section{Dataset Statistics and Analysis}
We follow a rough ratio of 8:1:1 to split our dataset into 8,566 for training, 1,095 for development, and 1,092 for testing. The train, dev, and test sets have no overlap tables.
We show the statistics of the dataset in Table~\ref{table:gen_stats} and the distributions of 7 logic types in Figure~\ref{fig:logic_dist}. Each table has 1-3 descriptions with different logic types. 
Since the logical forms present graph structure nature, we analyze the complexity of the logical forms based on the number of nodes in the graph, regarding the number of function nodes (\texttt{count}, \texttt{max}, etc.) and the number of all nodes (both function nodes and text nodes), respectively. As shown in Figure~\ref{fig:node_num}, the logical forms in \textsc{Logic2Text} have a minimum of 5 nodes and maximum over 14 nodes. For different logic types, \texttt{comparative} has the most number of nodes, because it involves the selection and operation for two table rows. \texttt{superlative}, \texttt{ordinal}, and  \texttt{unqiue} primarily focus on one table row, sometimes with the scope being a subset of all table rows, which makes the logical forms more complex. \texttt{count}, \texttt{majority}, and \texttt{aggregation} are summarization based logic types on multiple table rows. They are the three relatively simpler ones in terms of logical form structures. Figure~\ref{fig:structure} gives the logical form structures for 3 example logic types. 

%% file: 05-experiments.tex
\section{Experiments}
In this section we first describe the baseline models of our dataset in~\S\ref{exp:baselines}; Then we conduct experiments in fully-supervised setting~\S\ref{exp:fulldata}; We demonstrate the importance of the logical form in~\S\ref{exp:logical_form} and perform ablation studies in~\S\ref{exp:ablation}; At last we carry out experiments under few-shot setting~\S\ref{exp:fewshot}.

\subsection{Baseline Models}
\label{exp:baselines}
Apart from the logical forms serving as the primary input to the generation model, the table information is also crucial to provide context information. Following human's order to comprehend the table and produce descriptions, the input $C$ is formulated as the sequence of table captions, table headers, table content, and the logical form. The goal is to generate a sequence $w$ that maximize $P(w \mid C)$:
\begin{equation}
\begin{aligned}
    w = argmax \prod P(w_t \mid w_{0:t-1}, C)
\end{aligned}
\end{equation}

We employ the following models as our baselines for \textsc{Logic2Text}:

\textbf{Template} We manually craft generation templates for each logic type based on the logical form. 

\textbf{Seq2seq+att} We employ the seq2seq with an attention model from ~\cite{DBLP:journals/corr/BahdanauCB14}. The input sequence is formulated as the concatenation of the table caption, table headers, the linearized table content, and the linearized logical form.

\textbf{Pointer generator}~\cite{DBLP:conf/acl/SeeLM17} adds the copy mechanism upon the seq2seq with an attention model, allowing the decoder to copy tokens from the input directly. Such a mechanism is known to be critical for fidelity-preserving generation with abundant entities, numbers, etc.  

\textbf{Graph2seq+copy} There is a line of research for graph neural network based encoders, such as~\cite{DBLP:conf/inlg/MarcheggianiP18, DBLP:conf/emnlp/XuWWFS18}, etc. We employ one representative model, Graph2seq~\cite{DBLP:conf/emnlp/XuWWFS18}, to encode the logical forms. The table caption and headers are first fed into a seq2seq, followed by the graph encoder for the logical form. We also add the copy mechanism to allow copying from the input. 

\textbf{Transformer+copy} The popular Transformer model~\cite{DBLP:conf/nips/VaswaniSPUJGKP17} has shown remarkable progress in many tasks including NLG. 
In addition to the original Transformer structure, we add the copy mechanism where the last hidden layer is used to calculate the attention score and the copy switch. We also add segment embeddings for different input components, similar as~\cite{DBLP:conf/naacl/DevlinCLT19}. 

\textbf{GPT-2} Generally, with Transformer based structures, recent large-scale pre-trained models have achieved new SOTA results in a wide range of NLP tasks. A typical workflow is to use the pre-trained model as initialization, then fine-tune the model on task-specific data. In this work, we employ the generative pre-training model, GPT-2~\cite{radford2019language}, as one of our baselines.

For all neural models we use Byte-Pair Encoding (BPE) ~\cite{DBLP:conf/acl/SennrichHB16a} and the subword vocabulary used in~\cite{radford2019language}. Refer to Appendix C for more implementation details.

\subsection{Fully-Supervised Setting}
\label{exp:fulldata}
For automatic evaluations, we employ BLEU-4\footnote{
Standard script NIST mteval-v13a.pl} (B-4), ROUGE-1, 2, 4, and L (F measure)\footnote{rouge-1.5.5.}, noted as R-1, R-2, R-4, and R-L. The results for all baselines are presented in Table~\ref{table:res_full}. 

For models without pre-training, the copy mechanism brings a significant improvement, comparing pointer-generator and seq2seq. This is because the descriptions in our dataset involve much factual information from the table and the logical form, e.g., entity names, and numbers. 
However, the pre-trained language model GPT-2 can mostly accurately produce these factual terms even without a copy mechanism, demonstrating the powerful prior knowledge obtained from large-scale pre-training. 

\begin{table}[htbp]
\small
\begin{center}
\resizebox{.48\textwidth}{!}{%
\begin{tabular}{lccccc}
\toprule
Models & B-4 & R-1 & R-2 & R-4 & R-L \\
\midrule

Template & 17.57 & 50.56 & 24.20 & 6.61 & 37.81\\
\midrule
Seq2seq+att & 12.46 & 36.22 & 15.91 & 4.49 & 31.03\\
Pointer generator & 24.03 & 56.23 & 30.51 & 10.78 & 46.85\\
Graph2seq+copy & 25.38 & 58.15 & 32.79 & 12.25 & 49.47\\
Transformer+copy & 26.42 & 58.77 & 33.05 & 12.83 & 49.01\\
GPT-2 & \textbf{31.44} & \textbf{64.16} & \textbf{39.48} & \textbf{17.46} & \textbf{53.99}\\
\bottomrule
\end{tabular}
}
\caption{Automatic evaluation results for all baseline models under fully-supervised setting.}
\label{table:res_full}
\end{center}
\end{table}

Compared to the pointer generator, which takes linearized logical form as input, Graph2seq+copy directly models the graph structure and gets a slight improvement.
The Transformer+copy model obtains better performance than the Graph2seq+copy model, as the Transformer architecture is indeed a graph neural network with self-attention as aggregation function over the neighbors and regards the input as a fully-connected graph. Recent works~\cite{DBLP:journals/corr/abs-1906-01698, DBLP:journals/corr/abs-2002-12327, DBLP:journals/corr/abs-2005-09123} have shown that Transformer-based structure can capture hierarchical syntactic structures and graph representations. 
The GPT-2 model obtains the best performance among all with a significantly larger improvement. As a pre-trained language model with the Transformer structure, it combines the strength of both structural modeling and language modeling prior. Some example generations are provided in Appendix E. 
\subsection*{Human Evaluation}
Automatic scores are not sufficient for precise evaluation of factual and logical correctness. Therefore we conduct human evaluations through (1) crowdsourcing on Amazon Mechanical Turkers (AMT), and (2) human expert evaluations. 

For human evaluations on AMT, we randomly sample 500 examples from each of the top best-performing methods (GPT-2 and Transformer+copy), and the gold references. The evaluations are conducted on two axes: \textit{factual correctness} and \textit{language fluency}. For factual correctness, we ask the workers to verify whether the description is factually supported by the table; For language fluency, we conduct pairwise comparisons between different methods. For both evaluations, we distribute each task to 3 workers to eliminate human variance. The evaluation results of language fluency and factual correctness are shown in Table~\ref{table:res_amt_lan} and the first row of Table~\ref{table:res_amt_fact}, respectively. For more details of the evaluation, check Appendix D. 
\begin{table}[htbp]
\small
\begin{center}
\resizebox{.48\textwidth}{!}{%
\begin{tabular}{lccc}
\toprule
 & Gold & GPT-2 & Transformer+copy \\
\midrule
\% factually correct & 98.1 & 82.4 & 65.1 \\
\% semantically correct & 92.0 & 73.0 & 43.0 \\
\bottomrule
\end{tabular}
}
\caption{Human evaluation results of factual correctness (first row) and semantic correctness (second row). }
\label{table:res_amt_fact}
\end{center}
\end{table}
\begin{table}[htbp]
\small
\begin{center}
\resizebox{.48\textwidth}{!}{%
\begin{tabular}{lccc}
\toprule
& \% win & \% loss & \% tie \\
\midrule
GPT-2 vs Gold & 35.6 & 43.3 & 21.1 \\
GPT-2 vs Transformer+copy & 54.0 & 25.3 & 20.7 \\
Gold vs Transformer+copy & 61.2 & 23.6 & 15.2 \\
\bottomrule
\end{tabular}
}
\caption{Human evaluation results of language fluency. }
\label{table:res_amt_lan}
\end{center}
\end{table}
To conduct a precise evaluation of semantic correctness, i.e., whether the generation correctly matches the meaning of the logical form, we invite human experts (two computer science graduate students) to perform the evaluation. We sample 200 examples from each method and ask them to verify whether the description correctly presents the meaning of the logic form. Each example is examined by both students, and the decision is made after discussion. The second row of Table~\ref{table:res_amt_fact} shows the evaluation results. 

As we can observe from all evaluation results, the GPT-2 model gives big improvements on both fidelity preserving and language fluency, but there's still a gap, especially on semantic correctness. We believe our dataset can serve as a valuable resource posing such a challenge on high-fidelity generation with complex semantics. 
\subsection{Importance of the Logical Form}
\label{exp:logical_form}
We conduct experiments without using the logical form, i.e., to generate arbitrary logically correct descriptions solely based on the table, which is the task setting of~\cite{chen2020logic}. The generation is evaluated with all descriptions of the same table as multi-references, as in their setting. The best performing model of~\cite{chen2020logic} obtains a BLEU-4 score of 20.17 and a factual correctness rate of 20.2\% based on human evaluation of 500 samples. In contrast, the generations of our best -performing baseline can obtain a factual correctness rate of 82.4\% shown in Table~\ref{table:res_amt_fact}, which demonstrates the great importance of the logical form on high-fidelity generation. 
Note that the automatic scores are not directly comparable, since, in our task setting, each generation maps to a unique logical form and is evaluated with a single reference. 

\subsection{Component-Wise Ablation}
\label{exp:ablation}
\begin{table}[htbp]
\small
\begin{center}
\resizebox{.48\textwidth}{!}{%
\begin{tabular}{lccccc}
\toprule
Models & B-4 & R-1 & R-2 & R-4 & R-L \\
\midrule
GPT-2 & 31.44 & 64.16 & 39.48 & 17.46 & 53.99\\
\midrule
-w/o caption & 21.67 & 54.26 & 29.16 & 9.99 & 45.70\\
-w/o header & 29.86 & 62.98 & 38.46 & 16.64 & 52.57\\
-w/o content & 30.42 & 64.17 & 38.89 & 16.79 & 53.63\\
\bottomrule
\end{tabular}
}
\caption{Ablation study on other input components.}
\label{table:res_part}
\end{center}
\end{table}
We perform ablation studies on other input components: the table caption, header, and content, using the best-performing GPT-2 model.
As shown in Table~\ref{table:res_part}, both the table caption and header provide strong context information for generation, and the table content also brings a slight improvement. 
\subsection{Few-Shot Setting}
\label{exp:fewshot}
Considering that acquiring a large amount of (logical form, description) pairs in real-world cases is expensive, we also include a few-shot learning task for our dataset, where the model is only provided with hundreds of paired examples. Previous works have shown that the pre-trained language models obtain strong NLG performance even with a handful of fine-tuning instances~\cite{DBLP:journals/corr/abs-1904-09521}. Therefore we still use the best-performing GPT-2 model for this study. In our dataset, the amount of unseen logical form structures increases with the reduction of training instances. As shown in Table~\ref{table:res_few}, 
while there's still a gap with the fully-supervised result, the result with 1,000 training instances using GPT-2 is comparable to some other baselines with the full training data. This demonstrates the potential of incorporating generative pre-training for the few-shot learning task. 

\begin{table}[htbp]
\small
\begin{center}
\resizebox{.48\textwidth}{!}{%
\begin{tabular}{lccccc}
\toprule
\# of examples & B-4 & R-1 & R-2 & R-4 & R-L \\
\midrule
Full & 31.44 & 64.16 & 39.48 & 17.46 & 53.99\\
\midrule
100 & 17.09 & 48.26 & 23.52 & 7.47 & 38.74\\
200 & 19.98 & 51.99 & 27.02 & 9.42 & 41.86\\
500 & 23.04 & 56.64 & 30.99 & 11.35 & 46.86\\
1000 & 24.57 & 57.81 & 32.64 & 12.21 & 47.67\\
\bottomrule
\end{tabular}
}
\caption{Results for few-shot learning setting with 100, 200, 500, and 1000 training examples, using GPT-2. }
\label{table:res_few}
\end{center}
\end{table}
\vspace{-20pt}

%% file: 06-conclusion.tex
\section{Conclusion}
In this work, we formulate the problem of logical-level NLG as generation from logical forms in order to obtain controllable and high-fidelity generations. To this end, we propose a new dataset named \textsc{Logic2text}. 
There are some other potential future directions.
1) Human evaluations are precise but expensive. Our dataset can be used in the reverse direction to train a semantic parser, to assist parsing-based evaluations. 2) In this work, we primarily focus on the step to generate descriptions based on the logical form. 
Another potential future direction could be the content selections, i.e., how to select and organize the logical forms to construct a discourse plan based on user interests. 

%% file: 07-appendix.tex
\appendix

\section*{Appendix}
\subsection*{A. Logic Type Definitions \& Logical Form Annotation}

\subsection*{Logic Type Definitions}
\label{ap.a}
\begin{figure}[ht]
\centering
\includegraphics[width=0.48\textwidth]{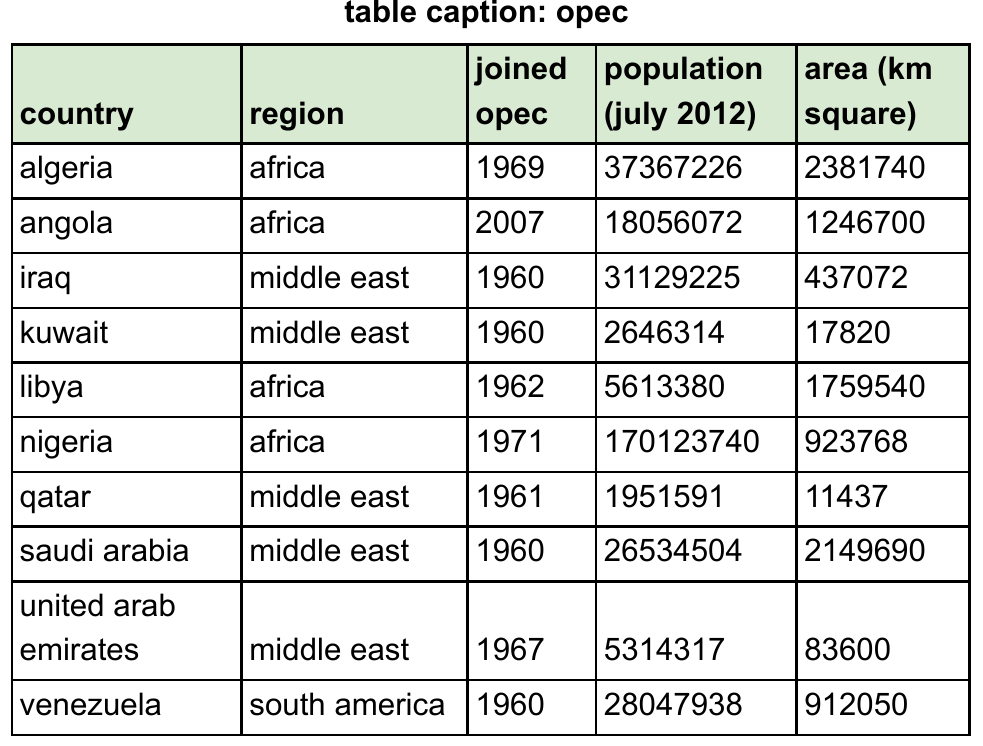}
\caption{Example table} 
\label{fig:app_eg0}
\end{figure}

We define all 7 logic types in our dataset and provide examples based on the following table in Figure~\ref{fig:app_eg0}.

\noindent\textbf{\texttt{Count}}: counting some rows in the table based on the values in one column, with the scope of all table rows or a subset of rows. 

\noindent\textbf{Example descriptions}: ``in opec 2012, there were 4 countries from africa.", ``in opec 2012, among the countries from africa, 2 of them joined after 1970.", etc.
\\

\noindent\textbf{\texttt{Superlative}}: Describing the maximum or minimum value in a column, with the scope of all table rows or a subset of rows. You may also talk about other columns on this row with the superlative value. 

\noindent\textbf{Example descriptions}: ``in opec in 2012, angola, from africa, was the latest country to join.", ``among the member countries in opec in 2012 from the middle east, qatar was the smallest in area.", etc.
\\

\noindent\textbf{\texttt{Ordinal}}: Describing the n-th maximum or minimum value in a column, with the scope of all table rows or a subset of rows. You may also talk about other columns on this row with the n-th maximum or minimum value. 

\noindent\textbf{Example descriptions}: ``in opec in 2012, qatar was the 5th country to join.", ``Among the africa member countries, algeria was the 2nd earliest to join.", etc.
\\

\noindent\textbf{\texttt{Comparative}}: Comparing two rows in the table, regarding their values in one column. You may also talk about other columns on these two rows. 

\noindent\textbf{Example descriptions}: ``in opec in 2012, libiya joined 2 years later than kuwait.", ``in opec in 2012, algeria, from africa, had a larger population than iraq from the middle east."
\\

\noindent\textbf{\texttt{Aggregation}}: Describing the sum or average value over a column, with the scope of all table rows or a subset of rows. 

\noindent\textbf{Example descriptions}: ``in opec 2012, the countries from africa had an average population of around 57,800,000.", etc.
\\

\noindent\textbf{\texttt{Unique}}: Describing one unique row, regarding one column, with the scope of all table rows or a subset of rows. You may also talk about other columns on this unique row. 

\noindent\textbf{Example descriptions}: ``in opec 2012, angola was the only country to join after 2000.", ``in 2012, among the member countries from africa, the only one to join opec after 2000 is angola.", etc.
\\

\noindent\textbf{\texttt{Majority}}: Describing the majority values (most or all) over one column, with the scope of all table rows or a subset of rows. 

\noindent\textbf{Example descriptions}: ``in opec 2012, most countries joined before 2000.", ``in opec 2012, all of the africa member countries had an area larger than 900,000.", etc.

\subsection*{Logical Form Annotation}
Here we provide the question sets for annotating each logical type.

\noindent\textbf{\texttt{Count}}:
(1). Choose whether the counting is performed on the scope of all table rows, or on a subset of all rows.
(2). Select the table column that the counting is performed on.
(3). Select the criterion, based on which we filter the table records to be counted. Here we consider the following criterion: "equal", "not equal", "less than", "less than or equal to", "greater than", "greater than or equal to", "fuzzily match", "all" (or "other" if none of the above is correct).
(4). Based on the selected criterion, write the value to be filtered for counting.
(5). Write down the result of the counting.

\noindent\textbf{\texttt{Superlative}}:
(1). Is the superlative action performed on the scope of all table rows, or on a subset of all rows?
(2). What is the table column that the superlative action is performed on?
(3). Is the superlative action taking the numerical maximum, or minimum value among the records?
(4). What is the table row containing this superlative value?
(5). On this row with the superlative value, what are the other column(s) mentioned? If not any other column is mentioned, write 'n/a'.
(6). Is this superlative value itself mentioned in the statement?

\noindent\textbf{\texttt{Aggregation}}:
(1). Choose whether the aggregation is performed on the scope of all table rows, or on a subset of all rows.
(2). Select the table column that the aggregation is performed on.
(3). What is the type of this aggregation, sum or average?
(4). What is the result of this aggregation?

\noindent\textbf{\texttt{Comparative}}:
(1). Which column is the statement comparing?
(2). What is the first row to be compared?
(3). What is the second row to be compared?
(4). What is the relationship comparing the records numerically in the first row with the second? (choose from "greater", "less", "equal", "not equal", "difference value", or "other" if not any of the above. Here we consider the relationship between actual numerical values between two records, NOT the relationship expressed in the statement )
(5). Is the compared records itself mentioned in the statement?
(6). What are the other column(s) of these two rows mentioned in the statement?

\noindent\textbf{\texttt{Majority}}:
(1). What is the scope of this majority?
(2). Which column the statement is describing?
(3). Is the statement describing all the records or most frequent records within the scope?
(4). Select the criterion, based on which we filter records to describe the majority. Here we consider the following criterion: "equal", "not equal", "less than", "less than or equal to", "greater than", "greater than or equal to", "fuzzily match" (or "other" if none of the above is correct).
(5). Based on the selected criterion, write the value to be filtered for describing the majority.

\noindent\textbf{\texttt{Ordinal}}:
(1). What is the scope that the ordinal description is performed on? (all rows or a subset of rows)
(2). What is the table column that the ordinal description is based on?
(3). Is the ordinal description based on a numerically max to min or min to max ranking of the column records?
(4). What is the order described in the statement, based on this ranking?
(5). What is the table row containing this n-th record ?
(6). On this row, what are the other column(s) mentioned? If not any other column is mentioned, write 'n/a'.
(7). Is this n-th record itself mentioned in the statement?

\noindent\textbf{\texttt{Unique}}:
(1). What is the scope of this statement describing unique row?
(2). What is this unique row?
(3). Write the table column that shows the uniqueness of this row
(4). Select the criterion, based on which we filter records in this column to find the unique row. Here we consider the following criterion: "equal", "not equal", "less than", "greater than", "fuzzily match" (or "other" if none of the above is correct).
(5). Based on the selected criterion, write the value to be filtered for the unqiue row.
(6). On this unique row, what are the other column(s) mentioned (except the column describing the scope)? If not any other column is mentioned, write 'n/a'.

\section*{B. Function Definitions}
Here we list the function definitions and descriptions for our logical form in table~\ref{table:app_func}. 
Note that since the tables in WikiTables are not standard database table, but semi-structured tables, the cell values are often not well-formatted with a lot of mixed strings and numbers, dates in different formats, etc. Therefore for some functions involving arithmetic operations on table cell values, we only specify a coarse ``object" type for the arguments, and then parse the numerical or date type values in the function implementations.
Refer to our released code for detailed implementations. 
\begin{table*}[ht]
\resizebox{\textwidth}{!}{%
\begin{tabular}{l|l|l|l}
\toprule
Name & Arguments & Output & Description \\
\midrule
count & view & number & returns the number of rows in the view \\
\midrule
only & view & bool & returns whether there is exactly one row in the view \\
\midrule
hop & row, header string & object & returns the value under the header column of the row \\
\midrule
and & bool, bool & bool & returns the boolean operation result of two arguments \\
\midrule
max/min/avg/sum & view, header string & number & returns the max/min/average/sum of the values under the header column \\
nth\_max/nth\_min & view, header string & number & returns the n-th max/n-th min of the values under the header column \\
\midrule
argmax/argmin & view, header string & row & returns the row with the max/min value in header column \\
nth\_argmax/nth\_argmin & view, header string & row & returns the row with the n-th max/min value in header column \\
\midrule
eq/not\_eq & object, object & bool & returns if the two arguments are equal \\
round\_eq & object, object & bool & returns if the two arguments are roughly equal under certain tolerance \\
greater/less & object, object & bool & returns if argument 1 is greater/less than argument 2 \\
\midrule
diff & object, object & object & returns the difference between two arguments \\
\midrule
filter\_eq/not\_eq & view, header string, object & view & returns the subview whose values under the header column is equal/not equal to argument 3 \\
filter\_greater/less & view, header string, object & view & returns the subview whose values under the header column is greater/less than argument 3 \\
filter\_greater\_eq /less\_eq & view, header string, object & view & returns the subview whose values under the header column is greater/less or equal than argument 3 \\
filter\_all & view, header string & view & returns the view itself for the case of describing the whole table \\
\midrule
all\_eq/not\_eq & view, header string, object & bool & returns whether all the values under the header column are equal/not equal to argument 3 \\
all\_greater/less & view, header string, object & bool & returns whether all the values under the header column are greater/less than argument 3 \\
all\_greater\_eq/less\_eq & view, header string, object & bool & returns whether all the values under the header column are greater/less or equal to argument 3 \\
\midrule
most\_eq/not\_eq & view, header string, object & bool & returns whether most of the values under the header column are equal/not equal to argument 3 \\
most\_greater/less & view, header string, object & bool & returns whether most of the values under the header column are greater/less than argument 3 \\
most\_greater\_eq/less\_eq & view, header string, object & bool & returns whether most of the values under the header column are greater/less or equal to argument 3 \\
\bottomrule
\end{tabular}
}
\caption{Function definitions}
\label{table:app_func}
\end{table*}

\section*{C. Model Implementation Details}
Here we provide some implementation details of the baseline models. 
\newline

\noindent\textbf{Template}
Some example templates are listed below. Texts in braces is optional depending on the logical form.

\noindent\textbf{\texttt{count}}: 

in [table\_caption], (among the ones whose [scope\_column] are [equal to/greater than/...] [scope\_value]), there are [result] ones whose [column\_name] are [equal to/greater than/...] [value] .

\noindent\textbf{\texttt{superlative}}: 

in [table\_caption], (among the ones whose [scope\_column] are [equal to/greater than/...] [scope\_value]), the [max/minimum] [column\_name] is [value].

in [table\_caption], (among the ones whose [scope\_column] are [equal to/greater than/...] [scope\_value]), [subject], with ([other\_col1] [other\_val];...), has the [max/minimum] [column\_name], ([value]).

\noindent\textbf{\texttt{ordinal}}: 

similar as \texttt{superlative}, replace max/minimum as n-th max/minimum.

\noindent\textbf{\texttt{comparative}}: 

in [table\_caption], [subject1] has [greater/less/...] [column\_name] than [subject2].

in [table\_caption], [subject1] has [diff\_value] [column\_name] [greater/less/...] than [subject2].

in [table\_caption], [subject1], with ([other\_col1] [other\_val];...), has [greater/less/...] [column\_name] than [subject2], with ([other\_col1] [other\_val];...).

\noindent\textbf{\texttt{unique}}: 

in [table\_caption], (among the ones whose [scope\_column] are [equal to/greater than/...] [scope\_value]), there is only one of them whose [column\_name] is [greater/less /...] than [value].

in [table\_caption], (among the ones whose [scope\_column] are [equal to/greater than/...] [scope\_value]), the only one whose [column\_name] is [greater/less/...] than [value] is for [subject], with ([other\_col1] [other\_val];...).

\noindent\textbf{\texttt{aggregation}}:

in [table\_caption], (among the ones whose [scope\_column] are [equal to/greater than/...] [scope\_value]), the [average/sum] of [column\_name] is [result].

\noindent\textbf{\texttt{majority}}:

in [table\_caption], (among the ones whose [scope\_column] are [equal to/greater than/...] [scope\_value]), [most/all] of them has [column\_name] [equal to/greater than/ ...] [majority\_value].
\newline

For all neural models we use Byte-Pair Encoding (BPE) ~\cite{DBLP:conf/acl/SennrichHB16a} and the subword vocabulary used in~\cite{radford2019language}. We use the pre-trained word embeddings from~\cite{radford2019language} and project to certain smaller dimensions (300) as the word embeddings. The batch size of all models are set to 32. The beam size is set to 3. As the table content only serves as context information for generation, to save GPU memory we set the maximum length of the table content as 200. The hyperparameters are chosen based on manual tuning regarding the BLEU score on the validation set. 
\newline

\noindent\textbf{Seq2seq+att \& pointer-generator} The learning rate is set to 0.001. For seq2seq, the training takes around 16000 gradient steps. For pointer generator, training takes around 5000 steps. 

\noindent\textbf{Graph2seq+copy} we reuse the code skeleton from the released code from~\cite{DBLP:conf/emnlp/XuWWFS18}. The table caption and header are first fed into a seq2seq, then the final hidden state is used to initialize the nodes of the graph encoder. When applying attention and copy, for graph nodes, we concatenate the token embedding and the embedding of its node as the embedding for the token. The learning rate is set to 0.0005. Training takes around 11000 steps. 

\noindent\textbf{Transformer+copy} we mostly follow the structure setting in the original Transformer model~\cite{DBLP:conf/nips/VaswaniSPUJGKP17}. We use 4 attention heads and 6 layers. The final hidden layer is used for calculating the  attention score and the copy switch. We also add the segment embeddings for different input components similar as~\cite{DBLP:conf/naacl/DevlinCLT19}. The learning rate is set to 0.0005. training takes around 32000 steps.

\noindent\textbf{GPT-2} We use the GPT-2 small 117M model from the released code and pre-trained model from~\cite{radford2019language}. Word embeddings are fixed during training. The learning rate is set to 0.0003. The training takes around 500 steps to converge.

All the experiments are run on GeForce GTX 1080Ti GPU. Table~\ref{table:res_val} shows the validation performance of different baselines. 

\begin{table}[htbp]
\small
\begin{center}
\resizebox{.48\textwidth}{!}{%
\begin{tabular}{lccccc}
\toprule
Models & B-4 & R-1 & R-2 & R-4 & R-L \\
\midrule
Template & 17.81 & 51.16 & 24.89 & 6.68 & 38.12\\
\midrule
Seq2seq+att & 12.26 & 35.44 & 15.68 & 4.81 & 30.36\\
Pointer generator & 25.43 & 57.35 & 31.97 & 12.33 & 48.11\\
Graph2seq+copy & 25.65 & 57.65 & 31.98 & 12.29 & 48.28\\
Transformer+copy & 27.20 & 59.70 & 34.06 & 14.03 & 48.71\\
GPT-2 & \textbf{32.98} & \textbf{64.86} & \textbf{40.02} & \textbf{18.38} & \textbf{54.59}\\
\bottomrule
\end{tabular}
}
\caption{Automatic evaluation results for validation set.}
\label{table:res_val}
\end{center}
\end{table}

\section*{D. Human Evaluation Details}
\noindent\textbf{Human Evaluations on AMT} We randomly sample 500 examples from the top two best performing methods (GPT-2 and Transformer+copy), and the gold references. The evaluations are conducted on two axes: \textit{factual correctness} and \textit{language fluency}. For factual correctness, we provide the workers with both the table and the description, and ask them to verify whether the description is factually correct based on the table. If the description contains too many grammar errors to be readable, the worker is instructed to select "incorrect". Minor grammar errors can be accepted, as long as the worker can understand the meanings. For language fluency, we conduct pairwise comparison between the three methods. For this evaluation we only present the pair of descriptions to the worker, and ask them to select a better one only based on language fluency (a better description should be fluent, coherent, and free of grammar errors), or select "Tied" if the two descriptions are of similar quality. 
For both evaluations we distribute each task to 3 workers to eliminate human variance. 

\noindent\textbf{Human Expert Evaluation} To conduct precise evaluation of semantic correctness, i.e., whether the generation correctly matches the meaning of the logical form, we invite human experts (two computer science graduate students) to perform the evaluation. We sample 200 examples from each method and ask them to verify whether the description correctly presents the meaning of the logic form, with neither insufficient nor redundant information. The description should also be fluent and free of grammar errors. Therefore this evaluation can be seen as a comprehensive evaluation of the generation quality. Each example is examined by both students and the decision is made after discussion.

\section*{E. Generation Examples}
\begin{figure*}[ht]
\centering
\includegraphics[width=0.96\textwidth]{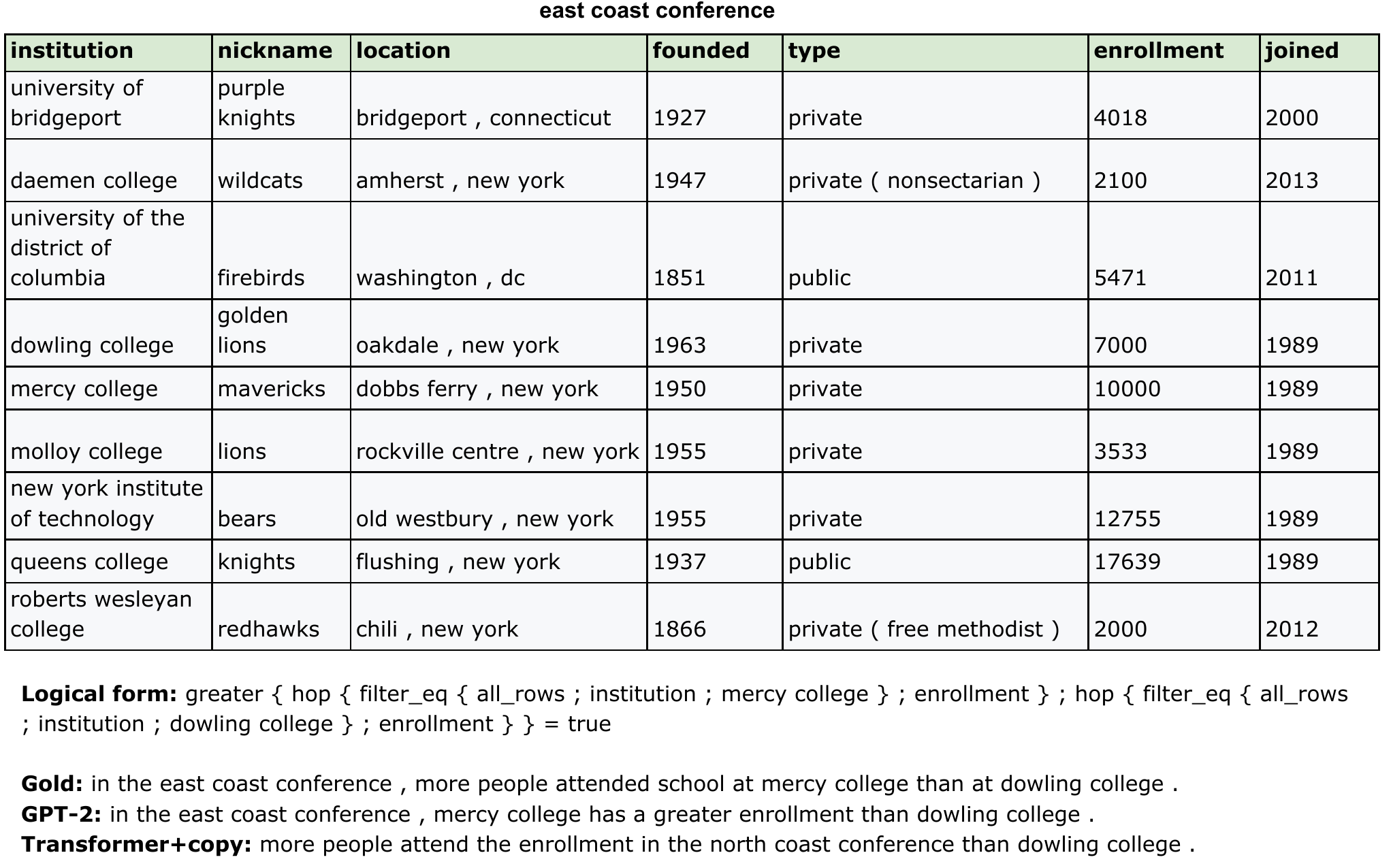}
\caption{Example generations.} 
\label{fig:app_eg1}
\end{figure*}

\begin{figure*}[ht]
\centering
\includegraphics[width=0.9\textwidth]{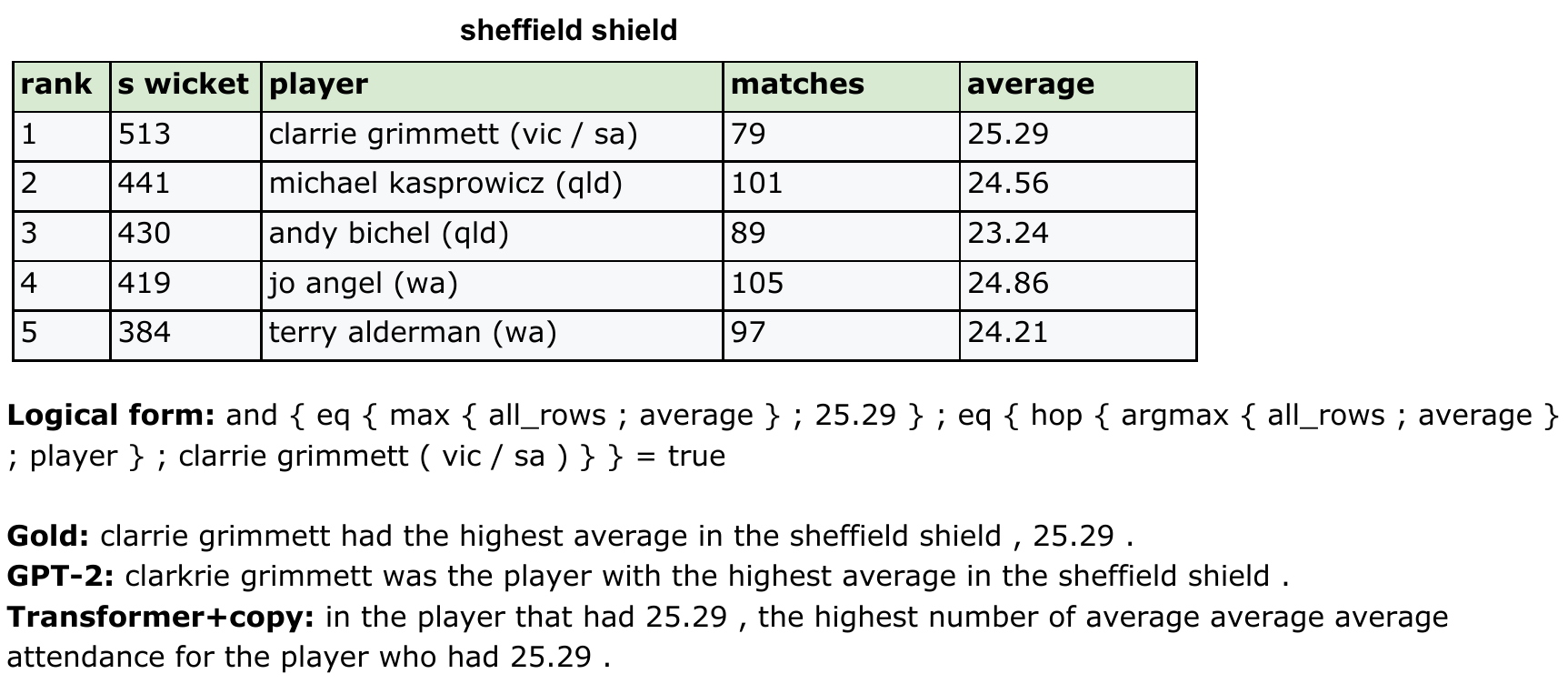}
\caption{Example generations.} 
\label{fig:app_eg2}
\end{figure*}

We provide 2 examples of generations in Figure~\ref{fig:app_eg1} and Figure~\ref{fig:app_eg2}.

